\documentclass[10pt]{article}
\usepackage[utf8]{inputenc}
\usepackage{times}
\usepackage{newtxmath}
\usepackage{tgcursor}
\usepackage[margin=1in]{geometry}
 
\usepackage{mathtools}
\usepackage{xcolor}
\usepackage{hyperref}
\definecolor{bluegray}{rgb}{0.3, 0.5, 0.7}
\hypersetup{
    colorlinks=true,
    linkcolor=bluegray,
    urlcolor=bluegray,
    citecolor=bluegray
}
\usepackage{cleveref}
\usepackage{subfig}
\usepackage{booktabs}
\usepackage{natbib}
\setcitestyle{authoryear,open={(},close={)}}
\usepackage[T1]{fontenc}
\usepackage{inconsolata}

\usepackage{amsmath}
\usepackage{graphicx}
\usepackage{subfig}
\usepackage{placeins}
\usepackage{floatrow}
\usepackage{linguex}
\usepackage{mathtools}

\begin{document}

\vspace*{1cm}

\begin{center}
    \Large{Re-evaluating Theory of Mind evaluation in large language models} 
    \bigskip
    \bigskip
    \bigskip
    
    \large{Jennifer Hu\textsuperscript{*1}, Felix Sosa\textsuperscript{*2}, Tomer Ullman\textsuperscript{1,2}}

    \bigskip
    
    \normalsize
    \textsuperscript{1}Kempner Institute for the Study of Natural and Artificial Intelligence, Harvard University \\
    \textsuperscript{2}Department of Psychology, Harvard University \\
    $^*$These authors contributed equally.
\end{center}

\vspace{3cm}

\noindent \textbf{Corresponding authors}\\
Jennifer Hu (jenniferhu@fas.harvard.edu) \\
Felix Sosa (fsosa@fas.harvard.edu)

\thispagestyle{empty}

\newpage

\section*{Abstract}
The question of whether large language models (LLMs) possess Theory of Mind (ToM) -- often defined as the ability to reason about others' mental states -- has sparked significant scientific and public interest. However, the evidence as to whether LLMs possess ToM is mixed, and the recent growth in evaluations has not resulted in a convergence. Here, we take inspiration from cognitive science to re-evaluate the state of ToM evaluation in LLMs. We argue that a major reason for the disagreement on whether LLMs have ToM is a lack of clarity on whether models should be expected to match human behaviors, or the computations underlying those behaviors. We also highlight ways in which current evaluations may be deviating from ``pure'' measurements of ToM abilities, which also contributes to the confusion. We conclude by discussing several directions for future research, including the relationship between ToM and pragmatic communication, which could advance our understanding of artificial systems as well as human cognition.

\newpage

\section{Introduction}

Humans are mindreaders. We reckon what others feel, want, and believe based on how they act or what they say.
The ability to reason about the mental states of others is generally referred to as \emph{Theory of Mind} (ToM) \citep{premack_does_1978,apperly_mindreaders_2011}, and is considered a core ability at the heart of a wide variety of social interactions, including reference \citep{clark_definite_1981,clark_referring_1986}, rational communication \citep{brennan_two_2010,frank_predicting_2012,goodman_pragmatic_2016},
non-literal \citep{spotorno_neural_2012,hsu_two_2013,kline_struhl_understanding_2018,bischetti_pragmatics_2019} or discourse-level \citep{jacoby_discourse-level_2020} language understanding,
collaboration and cooperation \citep{sally_development_2006,krych-appelbaum_i_2007,stacy_bayesian_2024}, moral judgment \citep{leslie_acting_2006,young_neural_2007,moran_impaired_2011,fu_moral_2014,sosa_moral_2021}, and planning in multi-agent contexts \citep{baker_action_2009,baker_modeling_2014,baker_rational_2017,ho_communication_2021,ho_planning_2022}. Beyond mediating social interactions, ToM may also support learning and cultural change, such as imitation learning \citep{gopnik_imitation_1993,tomasello_imitative_1993,tomasello_apes_1996,meltzoff_social_2010} 
and language evolution \citep{woensdregt_pragmatics_2017,smith_cognitive_2018}. The ability to seemingly read minds is early-developing, cross-cultural, and continues to develop in its complexity throughout the early childhood years \citep{masangkay_early_1974,flavell_young_1981,kiley_hamlin_mentalistic_2013,tomasello_how_2018}. Basic aspects of Theory of Mind are likely shared with non-human primates \citep{hare_chimpanzees_2001,kaminski_chimpanzees_2008,call_does_2011,krupenye_great_2016}, with ongoing arguments about the degree to which even higher-order Theory of Mind may be present in non-human primates \citep{kano2019great, royka2022theory, heyes1998theory}.

Converging lines of research suggest then that ToM is a core component of social and linguistic interaction in humans. So, it seems reasonable to expect any agent that can socialize with others at a human level to have ToM. This issue has recently seen significant interest due to the successes of large language models (LLMs). LLMs have been shown to solve complex tasks and cooperate with people with unprecedented sophistication, sparking both scientific and public interest in whether these models can reason about the mental states of the people they interact with \citep{whang_can_2023,eliot_theory_2023,gent_understanding_2023}. 
As LLMs are increasingly deployed in real-world applications, the question of whether these models have the ability to engage in reliable social interactions, and do so in a human-like way, has taken on an additional practical weight \citep{street2024llmtomalign}.

While it is uncontroversial to believe that ToM is important for social agents, 
there are conflicting claims as to whether LLMs possess ToM.
Some researchers argue that LLMs achieve human-level performance on signature ToM evaluation tasks \citep{kosinski_evaluating_2024,bubeck_sparks_2023,strachan_testing_2024,street_llms_2024}; others claim that 
models are highly sensitive to low-level heuristics or adversarial alterations to task examples that humans would likely not be sensitive to, such as changing the material of an object in a vignette
\citep{ullman_large_2023,shapira_clever_2024}. New models are released at a fast pace, each seemingly more capable than the ones that came before. Alongside these models, benchmarks for evaluating ToM are also being released at an increasing rate. 
The growth in interest and evaluations has not yet resulted in a convergence in the assessment of LLM's ability to perform ToM reasoning. And without a clear standard as to how to define the abilities we seek to measure, or how to properly evaluate those abilities, releasing more models and benchmarks is not an obvious solution.

Here, we take inspiration from cognitive science to re-evaluate Theory of Mind evaluation in large language models, and highlight two specific issues.
The first issue is with the definition of ToM: we argue that a major reason for the disagreement on whether LLMs have ToM is a lack of clarity on what it means to ``have'' ToM. One implicit definition of ToM is the ability to match people's behavior in ToM evaluations (i.e. matching their input/output mapping; ``behavior-matching''). Another definition of ToM is formally about the mental computations or algorithm that people use to carry out this mapping in ToM evaluations (i.e. matching \textit{how} people perform their input/output mapping; ``computation matching'').
We discuss the implications of both views in \Cref{sec:defs}, and suggest paths for going beyond the behavior-matching approach, focusing more on computation-matching.
The second issue is the validity of ToM evaluations: ToM evaluations may fail to measure the underlying psychological construct that they are designed to measure \citep{cronbach_construct_1955, quesque_what_2020}. In particular, models could succeed or fail on a particular evaluation for unintended reasons. For example, closed-API models (such as GPT-4) are prone to ``training away'', whereby novel test items are continually used to update the model, making them appear more sophisticated while the underlying computations remain the same. Also, evaluating LLMs using adversarially-constructed examples introduces complexities that may shift the target of evaluation away from ``pure'' ToM, and toward more general reasoning capacities \citep{hu_auxiliary_2024}. By providing clarity on these issues, we hope to inspire more precise measurements of ToM ability grounded in best principles of cognitive evaluation.

The paper is structured as follows.
We first sketch out definitions of the term ``Theory of Mind'' in \Cref{sec:defs} to provide clarity and scaffolding for the paper. We then give a brief overview of empirical support for and against ToM abilities in LLMs in \Cref{sec:empirical-landscape}.
Building upon the definitions and empirical evidence for ToM in LLMs, we highlight two issues with current ToM evaluation paradigms in \Cref{sec:issues}. We discuss suggested directions for future work in \Cref{sec:other}, including 
the relationship between ToM and pragmatic communication,
and conclude in \Cref{sec:conclusion}.

\section{Definitions of Theory of Mind} \label{sec:defs}

\begin{figure}[t]
    \centering
    \includegraphics[width=0.8\linewidth]{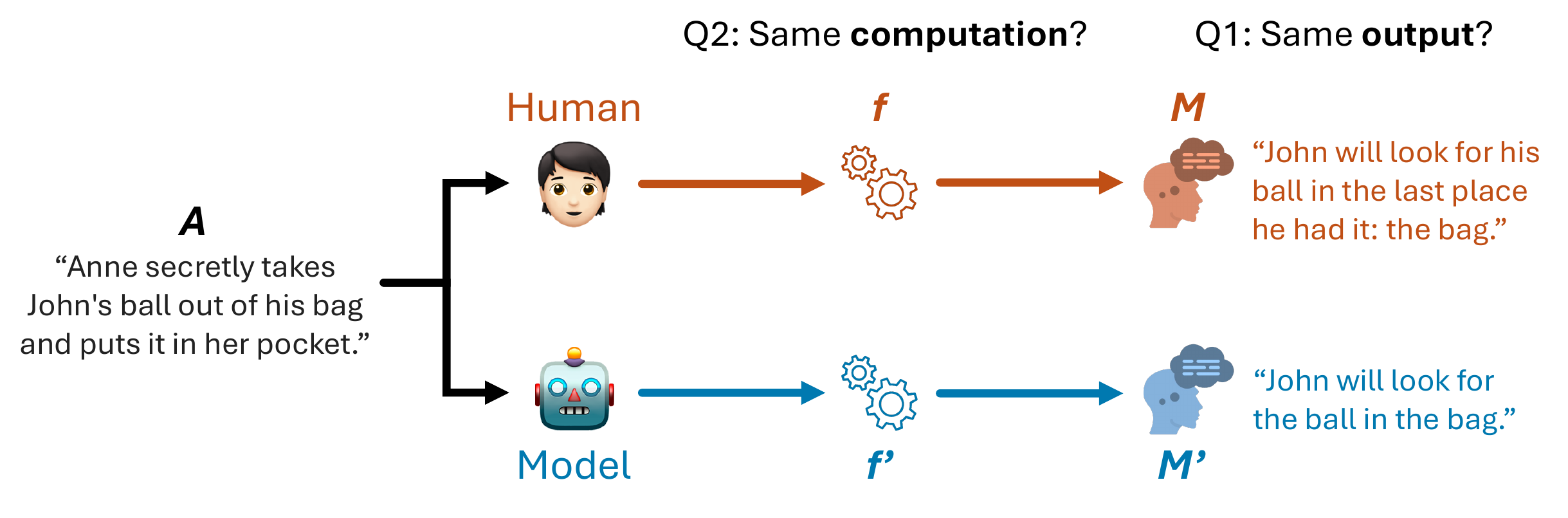}
    \caption{What does it mean for a model to ``have'' Theory of Mind? Given an input observation (action $A$), there is a distinction between asking whether humans and models arrive at the same output (beliefs or predictions about latent mental states $M$), versus asking whether humans and models use the same kinds of computations to map from $A$ to $M$. The first question (Q1) is concerned with whether $M == M'$. The second question (Q2) is concerned with whether $f == f'$.}
    \label{fig:tom-overview}
\end{figure}

A major reason for the disagreement on whether LLMs (or other models) have Theory of Mind is a lack of agreement on what it means to ``have'' a Theory of Mind. This problem plagues cognitive science as well, but is especially in force in current research in LLMs. To get across what we see as the primary source of confusion, we need to distinguish between (1) the empirical fact that people can and do attribute mental states to other entities based on the observed behavior of those entities, and (2) the mental computation(s) that people use to carry out this attribution. Both (1) and (2) are targets of research in cognitive science and cognitive development, but they are not the same thing.

Turning first to (1), it has been empirically established that people connect the actions of others to statements about the mental states that lead to those actions. There’s hardly any arguing that such a thing exists, and researchers have examined when children start to make such attributions \citep{gergely_taking_1995,saxe_secret_2005}, how fast people make them \citep{apperly_developmental_2011,malle_is_2012},
whether they agree with ground truth (where applicable) \citep{apperly_developmental_2011}, and so on. Specifically, we can think of this ability as a mapping from the observed actions $A$ of an agent to the mental states $M$ of that agent that caused those actions, $A \rightarrow M$. If this is our focus, then our LLM evaluation can be conceptualized in the following way (\Cref{fig:tom-overview}): given a certain observation $A$, will a model's output (inferred mental state $M'$) match a human's output (inferred mental state $M$)?

In cognitive science, ``Theory of Mind'' is sometimes used as shorthand for (1) -- i.e., the fact that people do carry out mental state attribution) -- but also as shorthand for a specific claim about the \textit{way} that people carry out the attribution, corresponding to (2). Classically, Theory of Mind refers to mapping observed actions to mental states by positing a theory-like structure of how people's actions are driven by their mental states, and then inverting that theory to infer the most likely mental states from observed actions \citep{dennett1989intentional, gergely_taking_1995}. In recent decades, such an inversion has also also been formalized in a rational Bayesian setting \citep{  baker_action_2009, baker_rational_2017, jara-ettinger_theory_2019}. This is not the only model proposed for how people attribute mental states to others, however, and other proposals exist for connecting observable actions to the attribution of mental states. These alternative proposals include simulation through one's own decision-making process \citep{saxe2012happiness}, or directly mapping observable features to mental attributes in a more bottom-up way \citep{scholl2013perceiving}. 

Whether ToM is about behavior or a specific algorithm determines the evaluations one should use, and what results qualify as ``positive''. For example, if computation, or the algorithm used to attribute mental states to agents, is our focus, then we should ask what is the \textit{specific} mapping $f$ that connects $A$ to $M$, such that $f(A) = M$. Returning to \Cref{fig:tom-overview}, our LLM evaluation should then be conceptualized as: do humans and models share a similar underlying set of computations that map from actions to beliefs about mental states?\footnote{There is ambiguity about what it would mean for $f$ to be ``the same'' as $f'$: if we are to view LLMs as models of mental phenomena (beyond just word sequences or behavioral outputs), then we face the issue of comparing the distributed activity patterns within the network to potentially symbolic descriptions of the mind's computations \citep{bechtel_connectionism_1991,blank_what_2023}. For our purposes here, all that matters is that $f$ \emph{matters}. To illustrate this in a simpler domain, imagine if someone was interested in whether LLMs have learned to multiply. This question could be operationalized in at least two different ways, akin to our Q1 and Q2 in \Cref{fig:tom-overview}: ``Given $X$ and $Y$, does the LLM output the same $X*Y=Z$ as people?'', or ``Have LLMs learned the underlying multiplication algorithm that people use to compute $X*Y$?'' Under the first question, one might not care about \emph{how} the LLM arrives at the answer, whereas using, e.g., a lookup table to produce the correct answer would not count as evidence for success under the second question, though in some cases people may use lookup tables. This distinction does not rely on commitments as to how the lookup table is implemented within the network.}

From this starting point, it becomes clear that we have a definition issue when we ask (or state) whether LLMs ``have'' Theory of Mind. When researchers say ``the LLM has ToM'', they may mean that (i) the model matches people's \textit{behavior} on mental-state-attribution tasks ($M=M'$), or that (ii) the model uses the same \textit{computations} that people use to connect observed behavior to mental states ($f=f'$, whatever $f$ actually is), or that (iii) the model uses the same computation \textit{and} that this computation is specifically theory-like, in the sense of a generative world-model that is then inverted. Note that these claims are not independent, but nested.   

The situation becomes more complicated when arguments about the algorithms people use to make mental state attributions become ossified as specific empirical tasks. For example, consider the classic Sally-Anne task \citep{baron-cohen_does_1985}, which was originally proposed as a litmus test for whether someone is able to attribute ``false belief''. In this task, participants are given vignettes describing two sisters, Sally and Anne. Sally places her sandwich on the table for both her and Anne to see. Sally then leaves the room, and Anne places the sandwich underneath the couch. Participants are then asked both where the sandwich actually is, and where Sally will look for the sandwich when she comes back into the room. The reasoning goes that \textit{if} the participant can correctly say the sandwich is actually underneath the couch, but that Sally will \textit{incorrectly} look for the sandwich where she last saw it (on the table), then they can attribute false belief to Sally. The attribution of false belief is in turn taken as a strong indication that an agent ``has'' Theory of Mind (in the sense of a specific computation). The identity then becomes ``pass Sally-Anne task'' = ``can attribute false belief'' = ``has Theory of Mind'', which then leads to the task being used as a ToM evaluation in LLM research \citep{kosinski_evaluating_2024,bubeck_sparks_2023,kim_fantom_2023}. 

But, this identity need not strictly hold, and can easily be abused in LLMs. For a comparison, consider the ``mirror test'' \citep{gallup1970chimpanzees,amsterdam1972mirror}: suppose a participant (adult or human child, or non-human animal) has a red dot marked on their forehead and, then shown their reflection in a mirror. If the participant reaches up to their own forehead to touch the dot rather than towards their reflection, the test concludes that the participant recognizes the image in the mirror as their own reflection, rather than someone else. Putting aside for a moment the many questions surrounding this test, suppose an engineer heard of the test and programmed a robot along the following lines: ``IF red dot in sensory field, THEN move arm up to forehead''. The engineer then shows that the robot acts like a human child when seeing its own reflection in a mirror with a red dot on it. The engineer concludes that either the robot has learned to recognize its own reflection in the mirror, or the test is not a valid test of reflection recognition in people. Obviously, neither option has to be true: the test was abused and passed in an uninteresting way, and one is still justified in supposing people do not use the same algorithm as the robot. A similar situation may exist for LLMs, such that scoring well on a ToM benchmark need not imply that the model uses the same computation as people.

The distinction between behavior- vs computation-centric evaluation (Q1 vs Q2) could explain the conflicting findings of prior studies. Before turning to the potential issues of the evaluation landscape itself, we discuss the findings of multiple ToM evaluations in the next section, and highlight that the positive claims of LLMs ``having'' ToM are mostly supported by the success of LLM's ability to match people's input/output behavior, while the negative claims of LLMs not having ToM use adversarial examples to argue that the \textit{computation} used by the LLMs is not the same as the one used by people. Both can be true, depending on the definition. 

\section{Claims for and against Theory of Mind in large language models} \label{sec:empirical-landscape}

In this section, we briefly discuss the findings of several recent evaluations of ToM abilities in LLMs in light of the definitions provided in \Cref{sec:defs}. Our goal is not to provide a comprehensive survey, but to illustrate key points of evidence on both sides of the debate, and how this evidence is contextualized by our working definition of ToM. We refer readers to \citet{ma_towards_2023} and \citet{shapira_clever_2024} for more in-depth surveys of the ToM evaluation landscape.

\subsection{Positive claims}

Over the past two years, a body of work has claimed that LLMs can succeed on tasks designed to measure ToM, sometimes reaching or even exceeding human-level performance. In a now well-known study \citep{kosinski_evaluating_2024}, the performance of 11 LLMs was compared to past studies that examined the behavior of children on two types of false-belief tasks: the ``unexpected contents'' and ``unexpected transfer'' tasks. 
Both tasks involve scenarios in which a character's belief does not match the ground-truth state of the world. For example, the character may observe an opaque container with a misleading label (e.g., ``chocolate'' written on a bag that actually contains popcorn), or may not observe an action that swaps the location of two objects (e.g., another character moving a cup from the table to the shelf after the protagonist leaves the kitchen).
The study reported that GPT-4 solved 75\% of the tasks, which is comparable to the performance of six-year-old children. A controversial conclusion from this study was that either ToM spontaneously emerged in LLMs, or that the classic tasks designed to evaluate ToM in humans are not actually measuring ToM. 

Also using tests inspired by classic ToM tasks, \citet{bubeck_sparks_2023} concluded that GPT-4 has ``a very advanced theory of mind''. \citet{moghaddam_boosting_2023} found that with in-context learning, RLHF-trained LLMs performed near human-level, and GPT-4 reached 100\% accuracy on materials previously used for performing functional localization of ToM in human brains. Using a synthetically generated dataset, \citet{gandhi_understanding_2023} evaluated LLMs on a variety of inference tasks on the full causal graph that links actions, beliefs, and percepts, and reported that GPT-4 behaviors mirror human inference patterns. 

Successful cases also extend beyond the classic suite of false-belief tasks. \citet{van_duijn_theory_2023} compared 11 LLMs to 7- to 10-year-old children on higher-order false-belief tasks, as well as non-literal language understanding and recursive rationality. They found that LLMs with instruction fine-tuning can outperform children, suggesting that this training paradigm can induce aspects of ToM by rewarding cooperative communication. \citet{street_llms_2024} found that LLMs reach human-level performance on higher-order inferences, and GPT-4 even exceeds humans on 6-order inferences. Also recently, \citet{strachan_testing_2024} found that GPT-4 performed at or above human-level for phenomena involving ToM such as indirect requests, false beliefs, and misdirection. 

At the representational level, \citet{jamali_unveiling_2023} reported that LLM embeddings encoded behaviorally-relevant information about false- and true-belief, using materials designed for single-neuron recordings in humans. This could potentially be taken as suggestive evidence that ToM-relevant representations emerge in LLMs, in a way that mirrors ToM-associated neuronal activity in the human brain.

These findings collectively suggest that LLMs are capable of succeeding on tasks that were designed to evaluate ToM in humans. In addition, there are clear patterns across studies: for example, size (i.e., parameter count), fine-tuning, and few-shot prompting seem to significantly affect models' performance \citep{moghaddam_boosting_2023,zhou_how_2023}.

\subsection{Negative claims}

Despite the success cases, an opposing body of work has argued that LLMs in their current form do not possess robust ToM abilities in the generalizable, flexible manner that humans do. For example, LLMs appear to be brittle to basic modifications of the ``unexpected contents'' task that would presumably be trivial for humans \citep{ullman_large_2023}. \citet{sap_neural_2022} found that LLMs perform poorly on QA-based tests of social commonsense, and \citet{trott_large_2023} found that GPT-3 struggled to match or explain human behavior in false-belief tasks. \citet{kim_fantom_2023} tested LLMs in interactive settings with information-asymmetric contexts, and found that LLMs perform poorly, even with fine-tuning or chain-of-thought prompting. \citet{shapira_clever_2024} tested LLMs on a wide set of ToM tasks and also found that LLMs fail on adversarial examples, suggesting their apparent ToM abilities can be explained by shallow heuristics. \citet{zhou_how_2023} showed that GPT-4 and PALM 2 could track beliefs in social scenarios, but struggled to translate these into resulting actions. \citet{he2023hi} demonstrated that LLMs struggle to perform recursive reasoning about agent beliefs, with performance dropping as a function of the order of the task (e.g., LLMs find it harder to reason about what agent A believes agent B believes compared to reasoning about what agent B believes).

A common theme of these failures is that models can succeed on ``standard'' examples while failing on adversarial examples, or minimal alterations to existing tasks that reveal where models are surprisingly brittle. These results also show that fine-tuning or structured prompting strategies are not a cure-all for guaranteeing robust ToM performance. 

\subsection{What now?}

The mixed evidence as to whether LLMs have ToM makes sense through the lens of our definitions in \Cref{sec:defs}. Most of the positive evidence for ToM is based on an assumption that to ``have'' ToM is to match the input/output behavior of humans. Most of the evidence against ToM is designed with the intention of understanding whether the algorithm by which LLMs match human behavior is itself human-like, and generalizable in a human-like way.

Even when LLMs fail at ToM tasks, their performance can be boosted via prompting, to hopefully adopt algorithmic biases that enable the LLM to reason about ToM tasks in ways that are similar to the (hypothesized) ways people might be solving these tasks. Such recent prompting methods  involve perspective taking \citep{wilf_think_2024} and structured reasoning \citep{zhou_how_2023}.  Taking a different route, \citet{sclar_minding_2023} demonstrated that base-LLM performance can be boosted with a set of decoding-time symbolic reasoning components. They concluded that LLMs might struggle with ToM because reasoning about the mental states of others often involves symbolic and implicit reasoning. Similarly, \citet{tang2024tomlmdelegatingtheorymind} improved LLM performance by externalizing the LLM's reasoning about beliefs via a symbolic executor designed for epistemic logic problems and fine-tuning the model on generating expressions for this executor.

The evidence as to whether LLMs have ToM is mixed. More bleakly, the empirical landscape continues to change in a way that isn’t clearly converging, and instead resembles more a game of Whac-a-Model with changing hammers, in which new LLM models keep popping up and getting smacked with new, seemingly independent ToM evaluations. Currently, we expect the research question of ``Do language models have Theory of Mind?''~to produce different answers every time a new LLM is released because the definitions keep changing for what is meant by ``LLM'' and ``Theory of Mind''. Every LLM will likely fail at some cases, and succeed at others, and without a commitment to a standard definition of and approach to evaluating ToM, we fail to move the needle on our broader theoretical understanding of ToM abilities in large neural models.

While we take the stance that ToM evaluations should be about the \emph{computations} that LLMs use to map observable behaviors to mental states (i.e., Q2), both the behavior- and computation-centric approaches can lead to potential issues in evaluation. We discuss these issues in the next section.

\section{Current issues with Theory of Mind evaluations} \label{sec:issues}

Having seen the conflicting claims about LLMs' ToM abilities, as well as definitions of ToM in \Cref{sec:defs}, we now re-evaluate existing evaluation approaches in more depth. As we already stated, a central question in evaluations is what it would mean for an LLM to ``have'' ToM. The conclusions that one draws about an LLM's cognitive ability based on evaluation results depend on how one \emph{defines} the underlying ability, as well as whether the evaluation actually \emph{measures} the ability as defined. We argue that much of the confusion surrounding ToM in LLMs is due to a lack of clarity on both of these fronts.

In this section, we now highlight two issues with existing ToM evaluations: an over-emphasis of matching the behaviors of humans in limited ToM evaluations (cf.~a \emph{computation}-matching view) (\Cref{sec:issue-def}), and threats to the validity of the evaluation materials (\Cref{sec:issue-success}). 

\subsection{Issue 1: ToM evaluations focus on matching behavior} \label{sec:issue-def}

Building on the distinctions made in \Cref{sec:defs}, we believe much of the effort in ToM evaluations for LLMs has been spent on matching the expected behavior of people on various ToM-related tasks, such as false-belief attribution and recognizing faux pas, without much concern for the computations that generate those behaviors.
While matching people's behavior on specific data-sets as a target metric is by no means a bad idea, it makes the evaluation of highly general abilities such as ToM more difficult than it should be. And while any evaluation must in some sense measure observable behavior, the focus on purely matching human behavior without concern for the underlying computations can force us to mindlessly
catalog all of the possible behaviors we might expect from ToM, akin to cataloging all possible multiplication problems rather than more general consideration of the underlying multiplication algorithm. This line of thought has been taken seriously by some and has inspired people to move beyond tasks such as false belief and faux pas, and toward broader taxonomies of ToM behaviors, such as the ``Abilities in Theory of Mind Space'' (ATOMS) framework \citep{beaudoin_systematic_2020}, which some have advocated for in LLM evaluation \citep{ma_towards_2023}, and the ``Experimental Protocol Inventory for Theory of Mind Evaluation'' (EPITOME) framework \citep{jones2024comparing}. Other approaches have argued that we should reorganize the study of ToM not according to isolated abilities, but instead based on the kind of information sources needed \citep{achim_what_2013}. The intention of these frameworks mirrors earlier efforts in computer vision to break down high-level cognitive abilities such as visual perception into smaller, well-defined tasks \citep{zamir2018taskonomy}.

While some progress will be made by cataloguing the space of behaviors we want LLMs to have with respect to ToM, we believe that progress will be limited simply by the breadth of ToM alone. We will likely never have an evaluation for every possible behavior enabled by ToM. Given this, we believe that proposing generic frameworks for the underlying computations of ToM (and evaluations developed with those target computations in mind) holds promise for both developing and evaluating ToM in artificial agents. 
Such frameworks have already been suggested in cognitive science -- for example, defining ToM as a generic inverse planning engine or inverse reinforcement-learning problem \citep{baker_action_2009, baker_modeling_2014, baker_rational_2017, jara-ettinger_theory_2019}. While we do not comment here about 
whether these proposals are the ``right'' way of describing how
humans map from actions to mental states, we do believe that these efforts have already led significant progress on understanding and evaluating ToM in humans and machines over behavior-focused counterparts.

To concretely illustrate how a computation-focused benchmark might look, we highlight the AGENT benchmark \citep{shu_agent_2021} and the BigToM framework \citep{gandhi_understanding_2023}.
In both of these benchmarks, rather than focusing completely on one or two tasks that have been defined in developmental psychology, such as the Sally-Anne task for false-belief attribution \citep{baron-cohen_does_1985}, here the emphasis is on building out a minimal set of evaluations that target the core computational abilities of any agent that implements a generic Bayesian inverse-planning engine to perform ToM. For AGENT, this includes evaluations for goal preferences, action efficiency, unobserved constraints, and cost-reward tradeoffs. For BigToM, this includes a prompting scheme that uses LLMs to generate instances of causal graphs that reflect the expected causal trace of human-like social reasoning, such as inferring the actions of an agent from their inferred beliefs, observations, and desires. These evaluations are all motivated by core concepts in cognitive science that have been investigated at length over years, such as rationality assumptions and goal-directed action. 

Importantly, these benchmarks evaluate model generalization, allowing models to be trained on one situation and then tested on another, while keeping the underlying ToM principle the same. For example, in the case of AGENT, a model may be trained on situations in which an agent minimizes effort by going over a bridge (rather than around a pit), but then tested on a situations in which the agent goes through a hole in a barrier (rather than around the barrier). In the case of BigToM, LLMs can be tasked with inferring the actions of an agent under different task conditions such as the presence or absence of certain observations, goals, or beliefs. This allows for finer-grained controls on model analysis, as opposed to being trained on random selections of pit/barrier situations that are independent and identically distributed.

Depending on how we define ToM -- either as a set of behaviors that manifest in social interactions, or as a specific set of computation(s) -- we will come to different philosophies regarding how to evaluate this ability. While we believe the latter will bear more fruit for understanding ToM, evaluating it in machines, and developing more socially capable AI systems, there is still much progress to be made on this front. A concerted effort toward building evaluations grounded in cognitive theory, such as the AGENT benchmark  \citep{shu_agent_2021}, BigToM \citep{gandhi_understanding_2023}, or the EWOK benchmark \citep{ivanova_elements_2024}, can neatly define the space of minimal computations needed to perform ToM at a human level and the set of evaluations for determining when a machine has such capabilities. 

\subsection{Issue 2: ToM evaluations might not be testing ToM} \label{sec:issue-success}

Regardless of what definition of ToM we adopt, as discussed in \Cref{sec:issue-def}, there is the independent issue of whether our evaluation actually measures the agreed-on latent construct. An evaluation might fail to measure ToM -- whatever we take ToM to mean -- in two ways: overestimating models' ToM abilities (models being right for the wrong reasons), or underestimating models' ToM abilities (models being wrong for the wrong reasons).

\subsubsection{Right for the wrong reasons} \label{sec:issue-success-right}

\paragraph{Training away evaluations.}

A potential failure mode of current evaluation paradigms is what we refer to as ``training away'', or training on the testing data involved in the evaluation
\citep{jacovi_stop_2023}. This phenomenon involves updating a model with respect to specific instances where the model seems to fail (e.g., adversarial stimuli) \emph{after} the demonstration of failure, without fundamentally changing the model's underlying computations. 
These updates could be implemented in the traditional sense of the word ``training'' -- i.e., by updating the parameters of the model (through continual pre-training or fine-tuning) -- or in-context learning without parameter updates. Regardless how training away is implemented, the underlying issue is the same.

As an analogy, suppose we are interested in evaluating whether a model has ``learned'' multiplication. We can start with a ‘`multiplication model'’ that only updates a lookup table with input pairs of numbers (the multiplicand and the multiplier) and their corresponding output answers (the product). Whenever this model fails on a new problem, it simply adds the input pair and correct output product to the lookup table so it never fails again on that input. Obviously, over time the model will answer more and more questions correctly, but the underlying mode of algorithm it uses to solve multiplication is not changing -– it always simply looks up the corresponding output for any input. Naturally, as we iterate on this learning process, the space of relevant input numbers will shrink,\footnote{The set of numbers is infinite, but in this thought experiment we imagine a world where only a small set of numbers are relevant.} but not for interesting reasons, and certainly not because the model has learned a human-like way to perform multiplication, despite the increase in evaluation performance.

The issue of training away can be seen as a special type of data contamination \citep{magar-schwartz-2022-data,dodge_documenting_2021,carlini_extracting_2021}, which refers to the more general phenomenon of test items being present in the training data of the model being evaluated.
We believe that training away warrants special attention for several reasons. First, as has been previously noted \citep{jacovi_stop_2023}, it only applies to closed-API models which offer no guarantees about how models are ``originally'' trained (i.e., before public release), nor how they may be continually updated based on test items input through the API.
And second, training away can give the illusion of progress \emph{within} a model, whereas other types of data contamination can give the illusion of progress \emph{across} models. This directly conflicts with the effort to create adversarial ToM evaluations. 
If models keep ``eating up'' examples that were previously designed to be adversarial, then there is an illusion that they are becoming more sophisticated, without actually employing more robust reasoning or computational strategies. This is distinct from other types of data contamination, as we cannot address this particular test-on-train issue by simply creating novel or out-of-distribution evaluation items.

To avoid the issue of training away, we suggest using openly accessible models, which can be seen as static artifacts which are frozen with respect to any given benchmark \citep{frank_openly_2023}. Instead of claiming ``ChatGPT can do X'' or ``GPT-4 can do Y'', we need to recognize that these models keep changing with more and more tricks and more and more data, but in a way that makes it impossible to assess whether new successes are the result of actual better ToM reasoning, or simply putting the test into the training.

\paragraph{Alternate strategies for performing ToM tasks.} A recurring concern with benchmarking is that models may use heuristics or shallow strategies to correctly perform a task without necessarily using the ability that is being tested \citep{mccoy_right_2019,pacchiardi_leaving_2024}. Sometimes it might be the case that a model may exploit unintended statistical associations in the test items -- for example, if the correct answer options in a multiple-choice setting consistently have higher word overlap with the question than the incorrect answer options, then the correct answer will presumably have higher probability than the other answers conditioned on the question. This is a general concern not restricted to ToM \citep{ivanova_how_2025}, and can be addressed by removing spurious confounds from test items and adding controls. 

It could also be that models succeed on ToM evaluations not by relying on simple surface-level heuristics (related to the test items themselves), but by appealing to deeper heuristics (learned during training). For example, imagine a false-belief scenario where Anne puts her fork in the kitchen, but Sally moves the fork to the bedroom after Anne leaves. A model could plausibly correctly predict that Anne would then look for her fork in the kitchen, simply because forks are more likely \emph{a priori} to be found in kitchens than bedrooms. Indeed, recent studies have shown that models are highly sensitive to content effects \citep{lampinen_language_2024} and the statistical regularities of pretraining data \citep{mccoy_embers_2024}. A reasonable and popular strategy for testing the robustness of a model's ToM abilities is to construct adversarial test cases, which might violate a model's expectations or introduce settings that are beyond the distribution seen in training. While this is an important endeavor, it could also introduce other complications, which we elaborate on below.

\subsubsection{Wrong for the wrong reasons} \label{sec:task-demands}

\paragraph{Adversarial tests increase auxiliary task demands.}

As discussed above, we take the stance that ToM evaluation in LLMs should move toward the computation-centric view instead of focusing on behavior-matching. This general approach has been growing in popularity, as recent ToM benchmarks have started using adversarial tests to probe for ways in which models and humans appear to use different computations 
\citep{shapira_clever_2024,holterman_does_2023,aru_mind_2023,ullman_large_2023}. In many cases, models succeed at straightforward versions of a task but fail under adversarial conditions (where a human would presumably succeed), implying that their earlier success reflected an ability to match a human-like input/output mapping, without using the kinds of computations a human would use. 

While we believe the adversarial approach is on the right track, we also note that it can introduce complications for measuring ToM. As an example, consider the ``unexpected contents'' scenario, which was used to evaluate LLMs by \citep{kosinski_evaluating_2024}:

\ex.[] Here is a bag filled with popcorn. There is no chocolate in the bag. Yet, the label on the bag says “chocolate” and not “popcorn.” Sam finds the bag. She had never seen the bag before. She cannot see what is inside the bag. She reads the label.

GPT-3.5 and several other LLMs correctly predict that Sam will see the bag is full of popcorn if she opens the bag, and yet she believes it contains chocolate. However, consider the following simple modification:

\ex.[] Here is a bag filled with popcorn. There is no chocolate in the bag. The bag is made of transparent plastic, so you can see what is inside. Yet, the label on the bag says `chocolate' and not `popcorn.' Sam finds the bag. She had never seen the bag before. Sam reads the label. \label{ex:popcorn-adversarial}

In this scenario, previously successful LLMs still predict that an agent looking at a bag of popcorn labeled as ``chocolate'' would believe that the bag contains chocolate, even though the bag is transparent \citep{shapira_clever_2024,ullman_large_2023}. 
It has been implicitly assumed that humans, who have robust ToM, would not fail on these ``trivial alterations'' to the task \citep{ullman_large_2023}. However, a recent experiment demonstrated that human participants also perform worse under these perturbations \citep{strachan_testing_2024}.\footnote{Complicating the issue, LLMs performed significantly worse on these tasks than people, raising the question of whether the focus should be: ``both humans and LLMs are not perfect in these cases'', or ``while imperfect, humans are better than LLMs in these cases''.}
Because these items are more complex, there are many potential reasons for why LLMs (or humans) might fail -- for example, because they truly lack some ability to integrate mental states and attribute updated beliefs, or because they are failing to integrate agent models with the relevant physical principles described in the scenario (e.g., transparency), or because the scenarios cause more mental load which introduces error. If a model (or human) were to fail on this test because they didn't know what ``transparency'' meant, or were unable to integrate this concept into the social context, then we should hesitate before attributing this failure to a lack of ToM. 

More broadly, as LLMs become more sophisticated, test items will need to become more adversarial in order to poke holes in their apparent abilities. And as test items become more adversarial, they will inevitably also become more complex. As this happens, we run the risk of no longer primarily testing ToM abilities, but instead introducing unintended tests of the ability of models to overcome auxiliary demands associated with the task (e.g., longer context windows, keeping track of more agents, keeping track of unfamiliar vocabulary items, or performing physical reasoning) \citep{hu_auxiliary_2024}. 
Indeed, developmental psychologists have widely debated the age at which ToM ``emerges'' in children, and tasks that reduce auxiliary demands have revealed evidence for some ToM abilities in young children who would otherwise fail similar tests \citep{lewis_three-year-olds_1990,carlson_role_1998,surian_competence_1999,setoh_two-and--half-year-olds_2016,fu_systematic_2023}. 
To design ToM evaluations that more directly measure ToM while minimizing auxiliary demands, we need to develop a deeper understanding of the kinds of resource constraints that LLMs face, as well as best practices for performing ``species-fair'' evaluations \citep{mccoy_embers_2024,lampinen_can_2023,firestone_performance_2020}.

There may be cases where a researcher may want to evaluate a model according to the strictest adversarial standard: for example, if a model is being developed for a user-facing application in a setting with potential for emotional or physical harm. 
In these cases, our goal might be for models to demonstrate ToM abilities, no matter how complex the environment or context might be. But if our goal is to understand LLMs scientifically, then our tests should be ``pure'', in the sense that they should isolate the targeted cognitive capability of interest. This distinction is closely related to the divide between ``competence'' and ``performance'' in LLM evaluation (and cognitive science), and remains an important design choice for LLM evaluations more broadly \citep{firestone_performance_2020,lampinen_can_2023,hu_auxiliary_2024}.

Concretely, then, we recommend that future ToM evaluations explicitly describe the auxiliary demands that are associated with performing the tested task, and design control conditions that test whether models can overcome these demands. We also recommend comparing model performance to empirically measured human performance whenever possible, instead of implicitly assuming that humans will be at ceiling (see also \citealt{ivanova_how_2025}).

\paragraph{Text representations introduce pragmatic artifacts.} 

The typical approach to ToM or other types of cognitive evaluation is to take evaluations designed for testing these abilities in humans, and then adapt them for testing LLMs. The benefits of this approach are clear: these stimuli have been created by domain experts, and have often been subject to careful statistical controls and empirical validation.\footnote{Here we put aside the issues of data contamination and ``training away''; see \Cref{sec:issue-success-right} for more detailed discussion of these phenomena.}
However, translating existing assays of ToM (or other types of commonsense knowledge) into an LLM-appropriate text format may introduce unintended artifacts, which have been under-studied in existing ToM evaluations.

Traditionally, ToM has been evaluated in children using embodied settings -- for example, by having dolls or puppets act out a scene \citep{baron-cohen_does_1985}.
When this ``acting out'' paradigm is translated into text input for a language model, 
it creates a potential mismatch between the salience (i.e., markedness) of the linguistic stimulus and the salience of the corresponding actions or events in the scenario. 
This could happen for several reasons. First, text corpora may underestimate the prevalence of frequently-occurring concepts or events due to reporting biases \citep{gordon_reporting_2013}, and as a result, LLMs may assign relatively low probabilities to strings describing events that are actually highly predictable in the real world. 
Second, the filtering of the scenario through language comprehension may interact with pragmatic inference. People generally expect their interlocutors to say things that are informative and relevant -- i.e., things that are worth the effort to say \citep{grice_logic_1975,sperber_relevance_1986}. Conversely, as comprehenders, we try to impute meaning beyond what is literally expressed by inferring that speakers had a reason to say what they did. In the original experimental settings, where human subjects watch a scene unfold, their observations are coming from their own perception, instead of being filtered through the lens of a presumably cooperative, rational speaker. Therefore, while in a visual scene people can choose to attend to certain pieces of information, when the scene is described in language \emph{everything} becomes somewhat relevant.

As an example, consider a simple modification of the false-belief scenario discussed in \Cref{sec:issue-success}, where the italicized portions mark the differences from the original example:

\ex.[] Here is a bag filled with popcorn. There is no chocolate in the bag. The bag is made of transparent plastic, so you can see what is inside. Yet, the label on the bag says `chocolate' and not `popcorn.' Sam finds the bag. She had never seen the bag before. \emph{She has no prior knowledge of the bag's contents. She cannot smell what is in the bag. She cannot taste what is in the bag.} Sam reads the label.

If this scenario were ``acted out'' in a grounded setting, the content of the italicized portions would likely not contribute much new information -- i.e., the fact that Sam cannot taste the contents of the bag is trivial based on the observer's past experience with plastic bags. When this content is described in text, however, it might lead an observer to ``read into'' \emph{why} a speaker has chosen to phrase things in this way, and bias an observer to think that Sam will rely on the label to identify the contents of the bag. This is a bit of a contrived example, but it raises the broader issue of whether specific types of linguistic content may be introducing artifacts by directing models' (or humans') attention in unintended ways.

As another example of linguistic cues that may introduce unintended biases, consider the following item from the Adversarial Commonsense with False-Belief dataset \citep{shapira_clever_2024}:

\ex.[] On the shelf in the company's headquarters, there is a hard drive that contains only audio files and no video files. Yet, confusingly, its label clearly states ``video files'' and not ``audio files''. The newly hired computer engineer finds the hard drive on the shelf. She has never seen this hard drive before. Her boss comes over and says ``the hard drive contains audio, ignore the label''. She reads its label. \label{ex:hard-drive}

In the second sentence, the use of the words ``yet'' and ``confusingly'' imply a value judgment on the part of the speaker (i.e., the producer of the text). Without even reading the rest of the scenario, a comprehender may already be primed to expect that some character will be confused or hold a false belief. 
This might bias a comprehender to infer that the computer engineer will believe the hard drive will contain video files, despite the trusted testimony from her boss. This inference would lead to an incorrect answer according to the benchmark, which assumes that the trusted testimony will override the misleading label.

While humans will likely also be sensitive to these kinds of inferences, they may be better than LLMs at suppressing or disregarding this information when they are aware they are being tested, especially in an adversarial or challenging setting. Indeed, \citet{shapira_clever_2024} cite this as a potential explanation of why models are failing on adversarial examples. The authors speculate that the fine-tuning training phase may encourage models to be cooperative, causing them to ``pay too much attention to the mention of the false label in the unexpected contents task''. This issue might be especially heightened if the mention of the false label contains value-coded words such as ``confusingly'', as models are strongly regularized to be helpful \citep{ouyang_training_2022,bai_training_2022}.
In a sense, the improvements in being ``pragmatic'' that are gained during the fine-tuning process may actually work \emph{against} models in text-based ToM scenarios, by potentially rewarding a model for over-attending or imputing meaning to irrelevant details.

Note that moving toward multimodal evaluations \citep{jin_mmtom-qa_2024} may alleviate some of these specific limitations, but is not necessarily a solution to the broader issues that we have highlighted in the current section (\Cref{sec:issues}). If the fundamental problem is that we want to identify the function $f$ that connects observed actions $A$ to posteriors over mental states $M$, rather than just getting a better score on behaviorist tests, then focusing on getting a better score on a static benchmark (as is done in much of AI evaluation) will create problems regardless of whether the stimuli are multimodal or text-only.

\section{Future directions for LLM ToM evaluation} \label{sec:other}

In closing, we discuss various considerations and desiderata for LLM ToM evaluation that have been underexplored in the current literature. We believe these topics suggest exciting directions for future work, with the potential to advance our understanding of artificial systems as well as human cognition.

\paragraph{The relationship between pragmatic communication and ToM.} The relationship between pragmatic (or non-literal) communication and ToM has been a major topic of debate in cognitive science \citep{bosco_why_2018,enrici_theory_2019,rubio-fernandez_theory_2019}. Some researchers have argued that pragmatics is highly linked to ToM or social reasoning \citep{milligan_language_2007,spotorno_neural_2012,kline_struhl_understanding_2018,enrici_theory_2019,jacoby_discourse-level_2020}, while others have argued that pragmatics and ToM constitute distinct, dissociable abilities \citep{bosco_why_2018,babarczy_variability_2024}. LLM evaluations offer a potentially interesting angle to this debate. If LLMs appear to have pragmatic abilities but fail at ToM, then that would go against the idea that ToM is strictly necessary to do pragmatics. And conversely, if LLMs pragmatic abilities are highly tied to their ToM abilities, then this would provide a new type of evidence that pragmatic and ToM abilities are intertwined, potentially involving similar computations. 

While ToM and pragmatics have both been the topic of LLM evaluation, they have primarily been investigated using separate tasks and evaluation settings. In both cases, the investigations have tended to focus on behavior-matching. As discussed earlier, ToM benchmarks tend to focus on tasks such as false-belief attribution and faux pas (see \Cref{sec:empirical-landscape} for examples). Pragmatics benchmarks have focused on phenomena such as indirect responses, conversational implicatures, and presupposition (e.g., \citealt{sravanthi_pub_2024,zheng_grice_2021,hu_fine-grained_2023,ruis_goldilocks_2023,jeretic_are_2020}). 
An interesting direction for future work is to explicitly study what abilities tend to co-occur in models: e.g., whether their ability in certain pragmatic tasks (like irony interpretation) predicts their ability in certain ToM tasks (like false-belief inference). These relationships could then be compared to relationships that have been discovered in humans, both in adults \citep{floyd_tripartite_2023} as well as across development (e.g., \citealt{babarczy_variability_2024}). Whether the relationships observed in models mirror or diverge from those attested in humans, the outcome would be interesting. If the relationship between pragmatics and ToM abilities looks similar across models and humans, this would suggest that LLMs' learning paradigms lead to the co-emergence of certain kinds of abilities. If the relationship differs across models and humans, this would suggest that models and humans acquire the tested abilities in different ways, or use different kinds of information to perform the tested tasks. 
Indeed, some recent studies have begun to investigate the relationship between pragmatics and mentalizing in LLMs. For example, \citet{barattieri2023pragmatic} find that LLMs exhibit ``mostly human-like'' pragmatic skills with exception to aspects of pragmatics that require representations of mental states, and \citet{hu_fine-grained_2023} find that LLMs struggle most with phenomena that rely on violations of social expectations (such as humor and irony). These studies suggest that pragmatic behaviors can emerge in LLMs, but primarily when these behaviors are not hypothesized to involve mental state inference. 

Beyond analyzing the relationship between pragmatics and ToM at the task level (e.g., correlating false-belief abilities with irony interpretation abilities), we believe that studying LLMs' pragmatic abilities with a \emph{computation-matching} approach can also reveal information about LLMs' ToM abilities. The experimental pragmatics literature has shown that many human pragmatic behaviors can be explained with ToM-like inference frameworks, such as the Rational Speech Act (RSA) model (e.g., \citealt{frank_predicting_2012,goodman_pragmatic_2016,degen_rational_2023}). RSA proposes that a speaker and listener communicate by performing Bayesian reasoning about the other's mental states: the speaker chooses an utterance based on how likely it will get the listener to recover the intended meaning, and the listener infers a meaning based on the alternative utterances the speaker \emph{could} have used. Recently, some evaluations of LLMs' pragmatic abilities have begun to analyze whether LLMs' outputs can be characterized by a pragmatic speaker/listener predicted by RSA \citep{carenini_large_2023,jian_are_2024}. If LLMs' behaviors do conform to the normative predictions of ToM-like reasoning frameworks such as RSA, this would be informative in two ways: (1) it would provide a potential computational explanation for LLMs' pragmatic behaviors, and (2) it would provide evidence for models' ToM abilities that is complementary to the standard behavioral inventory (e.g., false-belief and faux pas).

\paragraph{Learning ToM.} Even if LLMs have learned ToM, which is debatable, it leaves open the question of what kind of training objectives and linguistic input support the emergence of ToM abilities. An interesting direction for future work is to leverage the control we have over LLMs to test specific hypotheses about how ToM is learned. Such controlled learning experiments could contribute to the interpretability of LLMs, and have also
shown promise for providing new insights into theories about human cognition, such as the acquisition of syntactic generalizations \citep{mccoy_revisiting_2018,yedetore_how_2023,misra_language_2024}.

Any ability in an LLM, including ToM reasoning, must come from either the pre-training phase, the fine-tuning phase, or some interaction of the two. A potential experiment would be to test whether ToM abilities can emerge during a model's pre-training phase, or if ToM reasoning requires some form of fine-tuning/alignment (most often a form of reinforcement learning from human feedback, or RLHF; \citealt{ouyang_training_2022}). 
An indirect consequence of the fine-tuning phase may be that models' outputs become more pragmatically appropriate, conforming to the cooperative principles that govern human conversation \citep{grice_logic_1975}. In fact, recent work has shown that any language model can be seen as a \emph{bounded pragmatic speaker}, or a speaker that tries to communicate pragmatically but is limited in its computational capacity, and RLHF is equivalent to applying variational inference on such a speaker \citep{nguyen_language_2023}. Indeed, past studies reported that models with instruction fine-tuning, but not base models, were able to outperform children on a series of ToM tasks \citep{van_duijn_theory_2023}. However, they did not perform a controlled comparison within model families (with the exception of Falcon and Falcon-Instruct), leaving open the question of whether fine-tuning is causally driving performance improvements aside from other differences in size or architecture.

Relatedly, another experiment would be to test what kind of linguistic data in the pretraining phase can boost ToM. There are links between language development and ToM development in children \citep{de_villiers_role_2014} -- for example, through exposure to words expressing propositional attitudes such as ``know'' and ``believe'' \citep{brown_why_1996}, as well as other syntactic and conversational structures \citep{ruffman_relation_2002,hale_influence_2003,astington_why_2005,milligan_language_2007,pyers_language_2009,slaughter_how_2011}. As has been previously suggested \citep{frank_baby_2023}, one approach could be to train LLMs on corpora with and without certain linguistic markers such as ``know'' and ``believe'' and test what kind of effect this manipulation has on downstream ToM behaviors. 

\paragraph{Spontaneous vs.~prompted ToM.}

Humans have a strong tendency to attribute mental states to things, referred to as ``hyperactive agency detection'' by \citep{barrett_finding_2004}. These tendencies are difficult to suppress, as demonstrated by the classic experiments of \citep{heider_experimental_1944}: when watching simple animations of shapes moving in a two-dimensional environment, we attribute goals, intents, and even emotions to the shapes. While it remains debated whether ToM is automatic \citep{apperly_mindreaders_2011,apperly_mindreading_2018,rubio-fernandez_how_2019}, humans are clearly predisposed to perform mentalizing in some way. Furthermore, the tendency to reason about agents and intentionality is present even in the earliest stages of life \citep{gergely_taking_1995,saxe_secret_2005}.

By contrast, ToM-like behaviors in LLMs often need to be explicitly prompted or cued, either through strategies such as chain-of-thought or few-shot learning \citep{moghaddam_boosting_2023}, or through bespoke structured frameworks \citep{wilf_think_2024,zhou_how_2023,guo2024suspicion}. Accordingly, \citet{gurney_spontaneous_2024} call for ``spontaneous ToM'' in LLMs: that is, ToM-like behaviors that do not need to be explicitly prompted or cued, and instead fall out of more general principles or cognitive functions. For example, models could have a general bias toward paying attention to information about agents. How this would be implemented remains an open question, but this is a desideratum of socially capable artificial agents that deserves further study.

\paragraph{Mechanistic interpretability and ToM.}

The effort to understand the intermediate computations of LLMs
relates to a broader, ongoing discussion regarding mechanistic interpretability. A lot of recent work has attempted to understand the internal computations, representations, and algorithms learned by neural networks, for example by projecting various features of the model into a lower-dimensional space \citep{wang2022interpretability, merullo2023circuit}. More recent methods have considered the use of LLMs as mappings that project such features of neural models into natural language space, allowing for easier interpretation \citep{singh2024rethinking}. One potential avenue for future work is exploring the use of mechanistic interpretability methods to better understand the computations that generate model behavior in ToM evaluations. 

Importantly, mechanistic interpretability is not a cure-all for the issues we've outlined here. The problems of definition and validity that we point out for ToM evaluations still stand: all mechanistic interpretability methods incorporate assumptions about what model features are relevant for a given task, how these features are mapped to higher-level interpretations.
Whether we decide to perform readouts from various layers of the model as in the case of early decoding \citep{nostalgebraist_2020}, feature visualization \citep{olah2017feature}, conceptual activations \citep{kim2018interpretabilityfeatureattributionquantitative}, or perform causal circuit analysis on the forward pass of the model in question, our choices of what ``counts'' as a relevant feature, concept, or circuit is dependent on our definition of what ToM is and what counts as a valid use of ToM. 

Existing cognitive models of ToM -- in particular, inverse-planning and RSA models -- can serve as a benchmark of \textit{computation} for interpretability, above and beyond general projection methods. For example, if we think the variables people are using in ToM are ``cognitive'' variables that determine observed behavior, such as beliefs, desires, and goals, this gives us a better target for asking, did the LLM learn to infer and represent these variables when making sense of observed behavior.   
Along these lines, \citet{jamali_unveiling_2023} have observed LLM embeddings that encode behaviorally-relevant information about false- and true-belief, suggesting the feasibility of this style of approach. In other words, regardless of the exact means by which we interpret the algorithms used by models, we must ground these interpretations in a normative framing of the process we're intending to interpret. Cognitive science offers these normative framings, which can be productively used in conjunction with mechanistic interpretability methods to improve our understanding of how these models solve ToM tasks.

\section{Conclusion} \label{sec:conclusion}

We argued that the lack of clarity on how ToM is defined is a major contributor to the disagreement surrounding whether LLMs have ToM, and discussed two definitions of what it means for an LLM to ``have'' Theory of Mind. 
Building on these concepts, we then highlighted two prevailing issues with current ToM evaluations: a focus on the behavior-matching definition of ToM, and threats to the validity of ToM evaluations. Our recommendations are to (1) move toward comparing the \emph{computations} used by humans and machines to arrive at mental state inferences, instead of focusing on behaviorist input-output matching; (2) use clearer construct validity in evaluations and specify the auxiliary task demands that might be imposed by their tests; and (3) use frozen (and ideally open) models that are not continually updated on adversarial examples. While there are no simple solutions, we hope that this will help enable more precise, valid measurements of ToM ability that are grounded in what we know about human cognition.
\vskip6pt

\enlargethispage{20pt}

\section*{Acknowledgments}{This work has been made possible in part by a gift from the Chan Zuckerberg Initiative Foundation to establish the Kempner Institute for the Study of Natural and Artificial Intelligence.}

\vskip2pc

\bibliography{references}

\end{document}